\documentclass[acmengage,sigconf]{acmart}

\makeatletter
\def\@ACM@copyright@check@cc{} 
\makeatother

\usepackage{graphicx}
\usepackage{multirow}
\usepackage{amsmath}
\usepackage{relsize}
\usepackage{bbold}
\usepackage{caption} 
\usepackage{epsfig}
\usepackage{epstopdf}
\usepackage{algorithmic}
\usepackage{algorithm}
\usepackage{hyperref}
\usepackage{verbatim}
\usepackage{bm}

\usepackage{multirow}
\usepackage{pifont}
\usepackage{graphics}
\usepackage{adjustbox}
\usepackage{amsmath}
\usepackage{amsthm}

\usepackage{float}
\usepackage{pdflscape}
\usepackage{afterpage}
\usepackage{capt-of}
\usepackage{subfig}
\usepackage{bbm}
\usepackage[toc,page]{appendix}

\usepackage{url}
\makeatletter

\graphicspath{ {./images/} }

\usepackage[rightcaption]{sidecap}
\usepackage[colorinlistoftodos]{todonotes}
\usepackage{wrapfig}

\usepackage{booktabs}
\usepackage{placeins}
\usepackage[labelfont=bf]{caption}
\usepackage[normalem]{ulem}
\usepackage{color}
\usepackage{soul}
\usepackage{tikz} 
\usepackage{balance} 
\usepackage{paralist}
\usepackage[inline]{enumitem}

\def\@ACM@checkaffil{
  \if@ACM@instpresent\else
    \ClassWarningNoLine{\@classname}{No institution present for an affiliation}%
  \fi
  \if@ACM@citypresent\else
    \ClassWarningNoLine{\@classname}{No city present for an affiliation}%
  \fi
  \if@ACM@countrypresent\else
    \ClassWarningNoLine{\@classname}{No country present for an affiliation}%
  \fi
} 

\makeatother

\copyrightyear{2026}
\acmYear{2026}
\setcopyright{cc}
\setcctype{by}
\acmConference[CIKM '26]{Proceedings of the 35th ACM International Conference on Information and Knowledge Management}{November 07--11, 2026}{Rome, Italy}
\acmBooktitle{Proceedings of the 35th ACM International Conference on Information and Knowledge Management (CIKM '26), November 07--11, 2026, Rome, Italy}
\acmDOI{10.1145/3799682.3840133}
\acmISBN{979-8-4007-2539-5/2026/11}

\begin{document}
\title{Sequential Multimodal Evidence Optimization for Product Media Ranking in E-Commerce}

\author{Prasenjit Dey}
\affiliation{%
  \institution{Amazon.com Inc.}
  }
\email{prasendx@amazon.com}

\author{Frank McIntyre}
\affiliation{%
  \institution{Amazon.com Inc.}
  }
\email{flm@amazon.com}

\author{Arnab Sinha}
\affiliation{%
  \institution{Amazon.com Inc.}
  }
\email{arsinha@amazon.com}



\begin{abstract}
On modern e-commerce stores, customers consume ordered slates of heterogeneous product media—such as images, videos, and 3D renders—before making purchase decisions. Existing media-ranking systems often optimize myopic engagement proxies such as clicks or dwell time, even though product media assets are cooperative informational components of the same item that together help customers find the information they need through sequential interaction. We present Sequential Multimodal Evidence Optimization (SMEO), a two-stage utility-guided framework for customer-oriented media sequencing. SMEO first learns a trajectory utility model from consumed media prefixes to estimate how ordered evidence helps customers reach a purchase decision, while mitigating position-bias and variable-depth imbalance in logged data. Recognizing that customer attention is a limited resource, it then trains an autoregressive ranking policy with survival-weighted reward-to-go that prioritizes the most decision-relevant information early, so customers can find what they need with less effort. By decoupling utility learning from policy optimization, SMEO enables stable offline learning from biased logs and post-hoc media attribution without explicit media-level labels. Evaluated offline on large-scale e-commerce sessions using doubly robust off-policy estimation, SMEO improves estimated conversion by 5.5\% and helps customers reach a purchase decision with 15\% fewer swipes than existing baselines.

\end{abstract}

\ccsdesc[500]{Information systems~Learning to rank}
\ccsdesc[500]{Computing methodologies~Reinforcement learning}

\keywords{E-Commerce, Multimodal Learning, Policy Optimization, Ranking}
\maketitle

\section{Introduction}
\label{sec:intro}

\begin{figure}
\centering
\includegraphics[width=0.47\textwidth, height=6cm]{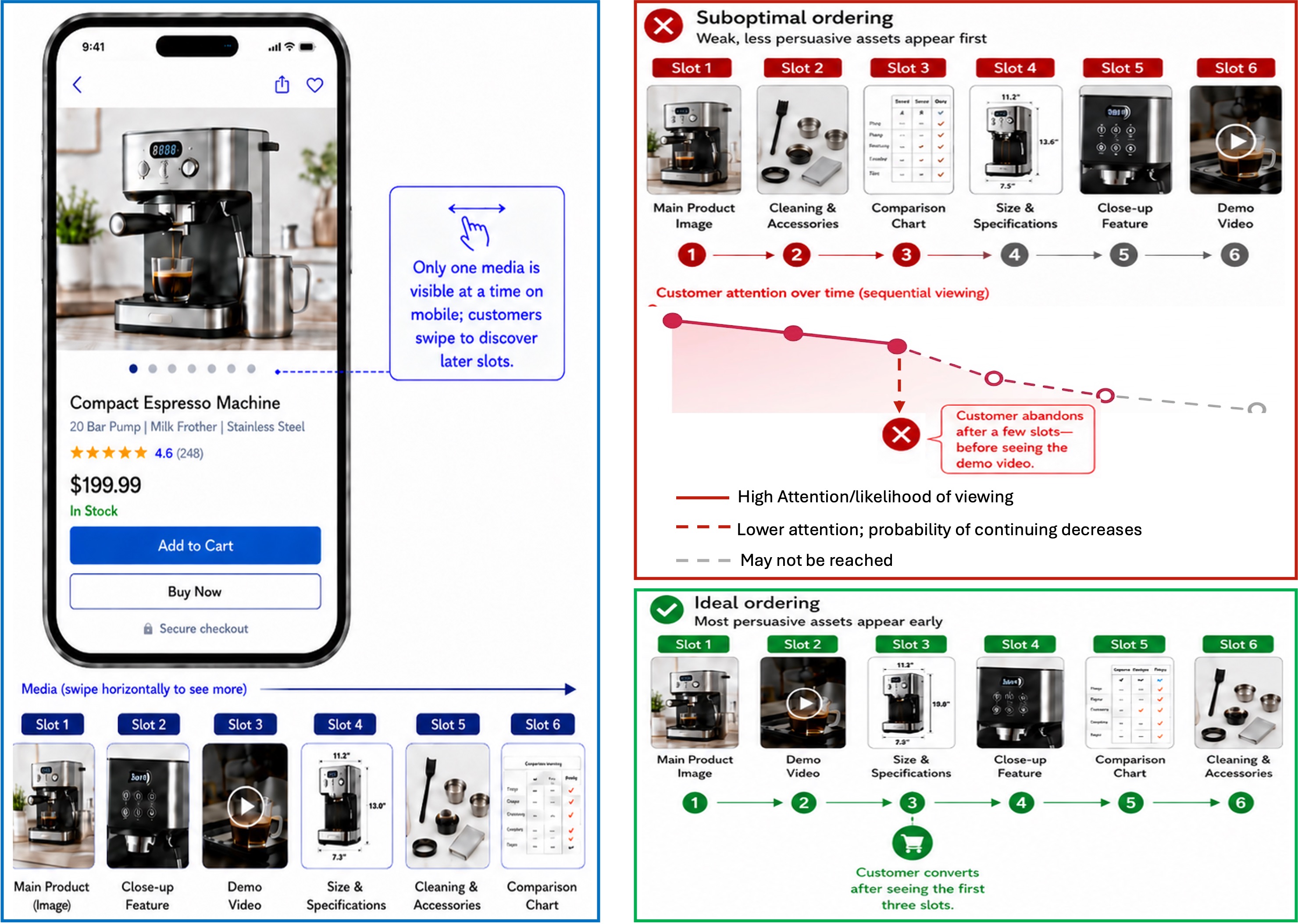}
\vspace{-2mm}
\caption{(Left) Product Detail page on Mobile. (Right) Suboptimal ordering may hide persuasive evidence until after customer drop-off, while ideal ordering front-loads decision-enabling media to support earlier conversion.}
\label{fig:dp}
\end{figure}

In modern e-commerce, the product detail page is the primary storefront. Customers are presented with a media carousel: an ordered slate of heterogeneous assets such as product images, lifestyle photographs, demonstration videos, interactive 3D renders, and virtual try-on experiences. Due to mobile viewport constraints, these assets are not shown simultaneously; customers must actively swipe or click to reveal later slots. If early assets fail to capture attention or convey product value, customers may abandon the page before discovering decision-relevant media placed deeper in the carousel, as illustrated in Figure~\ref{fig:dp}.

Because customers cannot physically inspect products online, purchase decisions depend on sequentially accumulating evidence from available media. This aligns with evidence accumulation models in cognitive science and decision theory~\cite{ratcliff2008diffusion}: customers consume informational signals until they either reach sufficient confidence to purchase or the interaction cost outweighs their interest. Therefore, learning the optimal ordering of multimodal product media to help customers find the information they need with less effort is a critical machine learning problem for e-commerce.

Prior studies show that high-quality product media improves conversion~\cite{nfinite2023visuals}. Despite its criticality, existing approaches in the literature rely on heuristic business rules or optimize for myopic engagement proxies (e.g., clicks, dwell time) that fail to guarantee terminal conversion \cite{zhou2018deep}. For instance an eye-catching image may generate clicks, but fail to provide the functional specifications a customer needs to make an informed purchase decision. Furthermore, Learning-to-Rank (LTR) frameworks \cite{liu2009learning, burges2010ranknet} inherently assume items are independent, competing alternatives. In reality, product page media are \textit{cooperative informational components}. The persuasive utility of a technical specification image is profoundly modulated by whether the customer first watched a lifestyle video establishing the product's value proposition.

Because cooperative media ranking introduces a fundamentally different learning problem, recent sequential slate optimizers like Seq2Slate \cite{bello2019seq2slate} and listwise evaluators \cite{pei2019personalized} prove inadequate. These methods expect dense, step-wise engagement over competing catalog items. Product media, by contrast, operates under severe reward sparsity where multiple assets jointly shape a single terminal purchase. Because media-level attribution is fundamentally unobserved in logs, standard methods naively assign equal credit to all engaged assets or rely on engagement proxies \cite{kang2018self,zhou2019deep}. This attribution crisis is compounded by \textit{endogenous truncation}—users often abandon sessions early, causing naive sequence models to misattribute terminal outcomes to unobserved media deep in the slate. Finally, historical default policies create substantial position bias \cite{zhao2019recommending}, confounding offline learning.
These properties---cooperative assets, unobserved attribution, endogenous truncation, and strong position bias---make product media ranking a distinct problem. To address this, we present \textbf{Sequential Multimodal Evidence Optimization (SMEO)}, a two-stage offline framework that shifts the objective from ranking independent media items to optimizing the sequential accumulation of persuasive product evidence.

SMEO first learns a trajectory utility model from logged data using the consumed prefix---the media actually observed before customer drop-off---while using position-aware utility learning and survival-based depth balancing to reduce bias from historical media placement and variable stopping. It then freezes the utility model and trains an autoregressive pointer network~\cite{vinyals2015pointer} using survival-weighted reward-to-go optimization,
where marginal utility gains are weighted by the probability that customers reach each position. This encourages the policy to surface the most decision-relevant media early, helping customers find what they need in fewer swipes.


The main contributions of this work are as follows:
\begin{enumerate}[leftmargin=*]
    \item \textbf{Novel Problem Formulation:} We formulate product media ranking as sequential multimodal evidence optimization, where assets for the same product cooperatively influence purchase decisions under variable customer stopping depths.

    \item \textbf{Utility-Guided Sequencing:} We propose SMEO, a two-stage
framework that learns conversion utility from consumed media
prefixes under position-biased logs and trains an autoregressive
policy to front-load decision relevant media.

\item \textbf{Media Attribution:} The learnt utility model enables post-hoc contribution analysis via counterfactual masking, without requiring explicit media-level conversion labels.

\item \textbf{Large Scale Offline Evaluation:} Evaluated with rigorous off-policy estimation, SMEO improves DR-CVR~\cite{dudik2011doubly} by 5.5\% and reduces swipes-to-conversion by 15\% over the existing baseline.

\end{enumerate}

\section{Related Work}
\label{sec:related}

\textbf{Sequential Recommendation and Slate Optimization:}
Sequential recommenders such as DIN~\cite{zhou2018deep}, DIEN~\cite{zhou2019deep}, and SASRec~\cite{kang2018self} model user histories to predict future item interactions. Slate-ranking methods, including DLCM~\cite{ai2018learning} and Seq2Slate~\cite{bello2019seq2slate}, capture intra-list dependencies via listwise scoring~\cite{cao2007learning} or autoregressive generation~\cite{jiang2018beyond,deffayet2023generative}. 
However, these target catalog recommendation or click modeling, where items are competing alternatives and feedback is typically item-level engagement. In contrast, SMEO models cooperative media assets for a single product under sparse terminal label, optimizing sequentially viewed media rather than next-item engagement or full-slate relevance.

\noindent \textbf{Credit Assignment and Counterfactual LTR:}
Attributing delayed outcomes to sequential interactions relates to Multi-Touch Attribution (MTA)~\cite{berman2018beyond,ren2018learning}, which assigns conversion credit across marketing touchpoints but does not produce intra-page media-ordering policies. Counterfactual Learning to Rank (CLTR)~\cite{joachims2017unbiased,swaminathan2015counterfactual,oosterhuis2022doubly} addresses exposure bias using propensity estimators, with extensions to sequential settings~\cite{xiao2023towards,wang2022unbiased}. SMEO is complementary: it does not claim full item-position identification, but reduces position-effect leakage via PAL\cite{zhao2019recommending} and handles endogenous carousel abandonment with survival-based depth balancing.

\noindent \textbf{Multimodal Ranking and Offline Policy Learning:}
E-commerce systems increasingly use visual and textual embeddings for ranking~\cite{tang2024multimodal,xu2024advancing}. These systems typically treat media as independent product features rather than an ordered sequence of persuasive evidence. SMEO draws on offline policy learning~\cite{chen2021decision,brandfonbrener2023offline} and RLHF-style reward-model separation~\cite{ouyang2022training,ziegler2019fine}, but adapts them to conversion-oriented media sequencing by decoupling trajectory utility learning from autoregressive ranking over biased offline logs.
\section{Problem Formulation}
\label{sec:problem_formulation}

Let $\mathcal{A}$ denote the set of products. Each product $a \in \mathcal{A}$ is associated with a multimodal media catalog
$\mathcal{M}^a = \{m_1^a, \ldots, m_{N_a}^a\}$, containing assets such as images, videos, and 3D renders. 
For a customer session $s$ on product $a_s$, a ranking policy presents an ordered media slate
$\pi^s \in \Pi(\mathcal{M}^{a_s})$, where $\Pi(\mathcal{M}^{a_s})$ denotes the set of feasible orderings of the product's media catalog. We write
$\pi^s=[m_{\sigma^s(1)}^{a_s},\ldots,m_{\sigma^s(N_{a_s})}^{a_s}]$,
where $\sigma^s(t)$ denotes the index of the media asset placed at position $t$.
Equivalently, $\pi_t^s=m_{\sigma^s(t)}^{a_s}$ denotes the asset at position $t$, and
$\pi_{\le t}^s=(\pi_1^s,\ldots,\pi_t^s)$ denotes the consumed prefix of length $t$.
When the session is clear from context, we omit the superscript $s$.

Customers consume the slate sequentially and may stop before reaching the end. We therefore observe only a consumed prefix
$\pi^s_{\leq T^s_{\mathrm{obs}}}$, where $T^s_{\mathrm{obs}}$ is the endogenous stopping depth. Each logged session is represented as
\[
s = \left(a_s, \pi^s_{\leq T^s_{\mathrm{obs}}}, y^s\right),
\]
where $y^s \in \{0,1\}$ denotes the terminal purchase outcome. 
Media assets placed at positions $t > T^s_{\mathrm{obs}}$ are not observed in the session and should not receive direct credit from the observed outcome.

Under this variable-depth, sparse-reward setting, we formulate product media ranking around two goals:

\begin{itemize}[leftmargin=*, topsep=2pt, itemsep=2pt]
    \item \textbf{Optimal Media Sequencing:} Learn a product-conditioned ranking policy that selects an ordering
    $\pi^*(a) \in \Pi(\mathcal{M}^a)$ to maximize expected terminal conversion under customer stopping behavior:
    \begin{equation}
        \pi^*(a)
        =
        \operatorname*{arg\,max}_{\pi \in \Pi(\mathcal{M}^a)}
        \mathbb{E}_{T \sim P_{\mathrm{stop}}(\cdot \mid a,\pi)}
        \left[
            P(y=1 \mid a, \pi_{\leq T})
        \right].
        \label{eq:ranking_objective}
    \end{equation}

    \item \textbf{Latent Media Contribution Estimation:} Estimate the marginal contribution of each media asset to terminal conversion, while accounting for the fact that media-level conversion labels are unobserved and contribution may depend on the surrounding media context and ordering.
\end{itemize}

This formulation differs from standard listwise ranking: the reward is sparse and terminal, the consumed trajectory is endogenously truncated, and media-level contribution is latent, context-dependent, and unobserved in logged data.


\begin{figure*}[t!]
\centering
\includegraphics[width=0.95\textwidth]{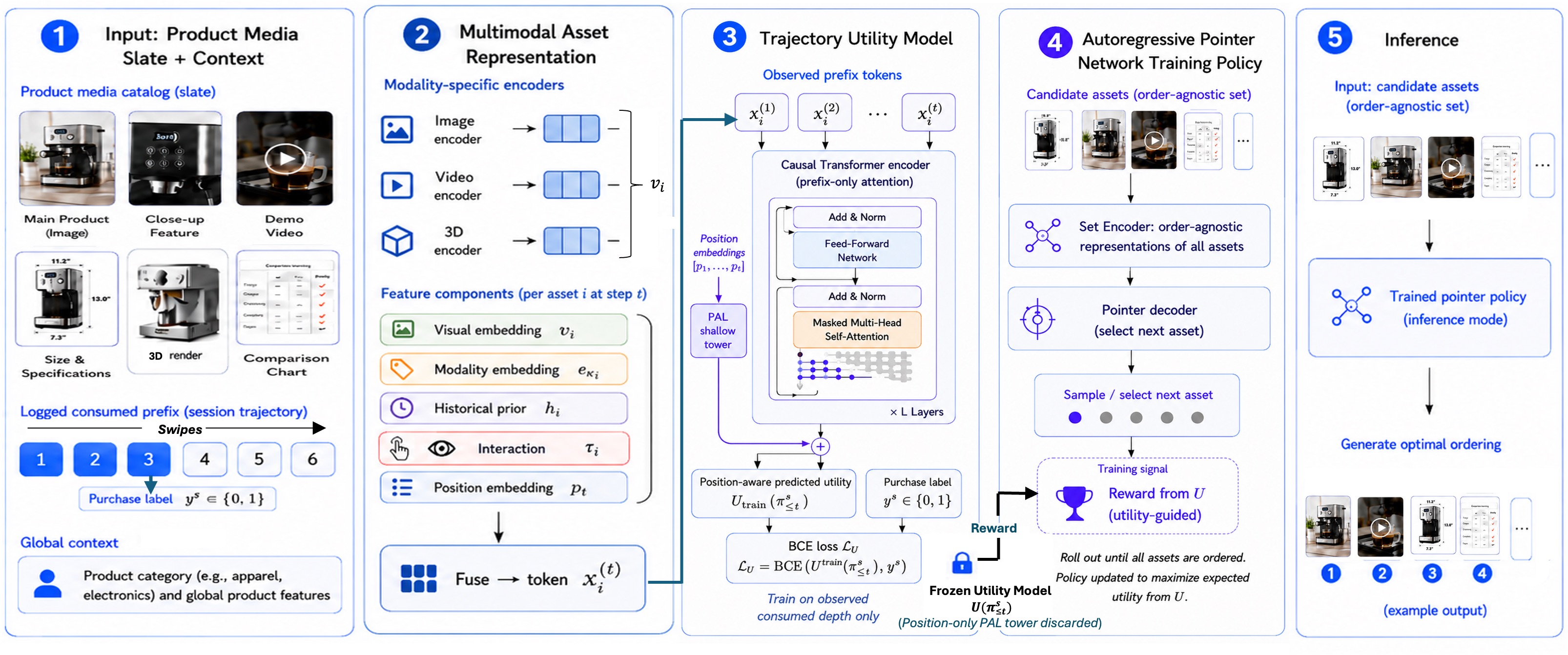}
\vspace{-2mm}
\caption{Complete end-to-end SMEO Framework}
\label{fig:wide_multimodal_pipeline}

\end{figure*}

\section{Proposed Methodology}
\label{sec:method}
Figure~\ref{fig:wide_multimodal_pipeline} summarises the SMEO framework.
We first construct multimodal asset representations
(Section~\ref{sec:representation}), then learn a trajectory utility model
$U$ from logged consumed prefixes under sparse terminal supervision
(Section~\ref{sec:utility}). We next train an autoregressive pointer
network policy that uses the frozen $U$ as a utility-guided reward model
to generate conversion-maximising media orderings
(Sections~\ref{sec:pointer}--\ref{sec:policy_optimization}).
Finally, we use the frozen utility model for offline media contribution
analysis (Section~\ref{sec:attribution}).

\subsection{Multimodal Asset Representation}
\label{sec:representation}

Each media asset $m_i^a \in \mathcal{M}^a$ is encoded into a unified
token $\mathbf{x}_i \in \mathbb{R}^d$ that fuses visual content, modality
type, and interaction signals.

\subsubsection{Visual Content Embedding}

Let $\kappa_i \in \mathcal{K} =
\{\texttt{img},\,\texttt{video},\,\texttt{3d}\}$
denote the modality type of asset $m_i^a$.
Visual content is encoded modality-conditionally:
\begin{equation}
  \mathbf{v}_i =
  \begin{cases}
    f_{\mathrm{ViT}}(m_i^a)  & \kappa_i = \texttt{img} \\
    f_{\mathrm{VMAE}}(m_i^a) & \kappa_i = \texttt{video} \\
    f_{\mathrm{Uni3D}}(m_i^a) & \kappa_i =\texttt{3d}
  \end{cases}
  \;\in\; \mathbb{R}^{d_v},
  \label{eq:visual_enc}
\end{equation}
where $f_{\mathrm{ViT}}$ is a frozen ViT-L/14 from CLIP~\cite{radford2021clip}, $f_{\mathrm{VMAE}}$ is a frozen VideoMAE encoder~\cite{tong2022videomae} that mean-pools frame-level features, and $f_{\mathrm{Uni3D}}$ is a frozen Uni3D foundation model~\cite{zhou2024uni3d} that encodes 3D point clouds and VTO meshes directly into the shared vision-language latent space, with all outputs projected to $d_v=256$.
A learnable modality-type embedding $\mathbf{e}_{\kappa_i} \in \mathbb{R}^{d_e}$ ($d_e=32$) captures distinct persuasive roles of each media type grounded in media function.

\subsubsection{Interaction Features and Train-Serve Decoupling}

Beyond visual semantics, we incorporate two numerical signal groups. First, a \textbf{historical prior vector} $\mathbf{h}_i \in \mathbb{R}^{d_h}$ captures aggregate asset-level statistics from past logs, such as average dwell time and click-through rates etc.; these priors are available at both training and inference time. Second, during training only, we extract in-session interaction signals $\boldsymbol{\tau}_i=[\log(1+d_i),\tilde{k}_i,z_i]\in\mathbb{R}^{d_\tau}$, where $d_i$ is normalized dwell time, $\tilde{k}_i$ is click count, and $z_i$ is zoom count. These signals provide direct gradient signal linking visual representations to observed engagement quality, but are unavailable at page-load inference.

To bridge this train--inference gap, we apply feature masking~\cite{srivastava2014dropout}: during training, $\boldsymbol{\tau}_i$ is replaced by a learned mask token $\mathbf{m}_\tau\in\mathbb{R}^{d_\tau}$ with probability $p_{\mathrm{drop}}=0.3$, and always masked at inference. This exposes the shared encoder to two regimes: interaction-visible examples provide engagement supervision, while masked examples force prediction from visual content and historical priors alone. The encoder thus learns engagement-correlated representations that remain calibrated without in-session interactions; the learned mask token also preserves activation statistics under layer normalization.

\subsubsection{Composite Token Representation}

The final token for asset $i$ at consumption position $t$ fuses visual, modality, historical, and in-session embeddings:
\begin{equation}
  \mathbf{x}_i^{(t)} \;=\;
  \mathrm{LN}\!\Bigl(
    W_v\mathbf{v}_i + W_e\mathbf{e}_{\kappa_i} + W_h\mathbf{h}_i
    + W_\tau\tilde{\boldsymbol{\tau}}_i + \mathbf{p}_t
  \Bigr)
  \;\in\;\mathbb{R}^d,
  \label{eq:token}
\end{equation}
where $\tilde{\boldsymbol{\tau}}_i \in \{\boldsymbol{\tau}_i, \mathbf{m}_\tau\}$ is the conditionally masked interaction vector, $W_{v,e,h,\tau}$ are learnable linear projections, $\mathbf{p}_t$ is a sinusoidal positional embedding~\cite{vaswani2017attention} capturing evidence-accumulation order within the slate, $\mathrm{LN}$ denotes layer normalisation, and $d=256$. These position-aware tokens form the observed prefix sequence consumed by the trajectory utility model.

\subsection{Trajectory Utility Model}
\label{sec:utility}

\subsubsection{Causal Sequence Encoder and Position-Aware Learning}

We parameterise the trajectory utility function as an $L$-layer Transformer encoder~\cite{vaswani2017attention} with \emph{causal} left-to-right self-attention masking. For a consumed prefix $\pi^s_{\leq t}$, the encoder produces contextualised hidden states representing cumulative evidential belief:
\begin{equation}
  \bigl[\mathbf{h}_1,\ldots,\mathbf{h}_t\bigr]
  =
  \mathrm{TransEnc}_{\mathrm{causal}}
  \!\Bigl(
  \bigl[\mathbf{x}_{\sigma(1)}^{(1)},\ldots,\mathbf{x}_{\sigma(t)}^{(t)}\bigr]
  \Bigr).
  \label{eq:encoder}
\end{equation}

To reduce leakage of systematic position effects into the content encoder, we employ a Position-Aware Learning (PAL) architecture~\cite{zhao2019recommending}. We decompose the utility logit into a content-driven component from the Transformer state and a position-only component from a shallow network that observes only positional embeddings:
\begin{align}
  s_{\mathrm{content}} &= \mathrm{MLP}_{\mathrm{content}}(\mathbf{h}_t), 
  s_{\mathrm{pos}} = \mathrm{MLP}_{\mathrm{pos}}([\mathbf{p}_1,\ldots,\mathbf{p}_t]).
\end{align}
During utility-model training, the prediction uses the sum of both logits: \qquad \quad
$
  U_{\mathrm{train}}(\pi_{\leq t}^s)
  =
  \operatorname{sigmoid}(s_{\mathrm{content}} + s_{\mathrm{pos}}).
$

The position-only tower acts as a nuisance pathway for systematic slot-depth effects, reducing position-bias leakage into the sequence encoder. During training, it receives only consumed-prefix positional embeddings, with $10\%$ dropout to limit over-reliance on slot features. During policy optimization and inference, the PAL tower is discarded, and the frozen reward model uses the PAL-adjusted content utility, as shown in Fig.~\ref{fig:wide_multimodal_pipeline}:
\begin{equation}
  U(\pi_{\leq t}^s)
  =
  \operatorname{sigmoid}(s_{\mathrm{content}})
  \in [0,1].
  \label{eq:utility_head}
\end{equation}
This mitigates position-bias leakage from biased logged data.

\subsubsection{Variable-Depth Supervision}
\label{sec:vd_supervision}

The utility model is trained on logged sessions
$\mathcal{D}=\{(a_s,\pi^s_{\leq T^s_{\mathrm{obs}}},y^s)\}_{s=1}^{|\mathcal{D}|}$.
The observed media evidence available to influence $y^s$ is restricted to the consumed prefix: assets beyond $T^s_{\mathrm{obs}}$ were not viewed and should not receive direct credit for the purchase outcome. Each session therefore supervises the model only at its observed consumed depth:
\begin{equation}
  \mathcal{L}_U
  =
  -\frac{1}{|\mathcal{D}|}
  \sum_{s\in\mathcal{D}}
  \hat{w}^s
  \Bigl[
    y^s\log U_{\mathrm{train}}\!\bigl(\pi^s_{\leq T^s_{\mathrm{obs}}}\bigr)
    +(1{-}y^s)\log\!\bigl(1-U_{\mathrm{train}}\!\bigl(\pi^s_{\leq T^s_{\mathrm{obs}}}\bigr)\bigr)
  \Bigr],
  \label{eq:utility_loss}
\end{equation}
where $\hat{w}^s$ is a trajectory balancing weight (Section~\ref{sec:ips}).

\noindent\textbf{Remark (why all-prefix supervision is incorrect).}
A natural alternative augments $\mathcal{L}_U$ with
$\mathrm{BCE}(U_{\mathrm{train}}(\pi^s_{\leq t}),y^s)$ for every $t<T^s_{\mathrm{obs}}$.
For example, in a session
\[
[\texttt{weak\_img},\,\texttt{weak\_img},\,\texttt{great\_video}]
\rightarrow y=1,
\]
supervising $U_{\mathrm{train}}(\pi_{\leq 1})$ toward $1$ can incorrectly credit the weak image for a conversion observed only after the later video was consumed. Equation \ref{eq:utility_loss} avoids this leakage by supervising each session only at its observed consumed depth.

\subsubsection{Stopping Depth Balancing}
\label{sec:ips}

Product detail page logs are dominated by shallow prefixes and relatively few deep prefixes because customers often stop before reaching later media, causing naive training to underfit deeper consumed trajectories. Unlike standard LTR, which corrects item-level examination bias for pointwise rewards~\cite{joachims2017unbiased}, our utility model predicts terminal conversion for an observed media trajectory. We therefore use survival-based depth reweighting to balance observations across stopping depths:
$$
  \hat{\rho}_t
  =
  \frac{\#\{s\in\mathcal{D}:T^s_{\mathrm{obs}}\ge t\}}{|\mathcal{D}|},
  \qquad
  \hat{w}^s
  =
  \min\!\left(
  \frac{1}{\hat{\rho}_{T^s_{\mathrm{obs}}}},
  W_{\max}
  \right),
  \quad W_{\max}=10.
  \label{eq:depth_weight}
$$
Here, $\hat{\rho}_t$ is the empirical probability of reaching depth $t$, and $\hat{w}^s$ upweights rare deep prefixes while clipping controls variance.

\subsection{Autoregressive Pointer Network Policy}
\label{sec:pointer}
After training, $U$ from Eq.~\ref{eq:utility_head} is frozen and used only to score candidate prefixes during policy optimization.
We parameterise the ranking policy $p_\theta$ as a pointer
network~\cite{vinyals2015pointer} --- an autoregressive model whose output vocabulary is the input set itself. Three properties motivate this choice over a standard decoder.
First, it generalises across products without retraining, operating on asset \emph{embeddings} rather than fixed item IDs. Second, autoregressive masked softmax guarantees each asset appears
exactly once in the output. Third, it scales naturally to variable slate sizes $N_a\in[1,N_{\max}]$.

\subsubsection{Set Encoder} 
The encoder produces order-agnostic representations of all assets:
$$
\bigl[\bar{\mathbf{h}}_1,\ldots,\bar{\mathbf{h}}_{N_a}\bigr] \;=\; \mathrm{TransEnc}_{\mathrm{full}} \!\bigl([\mathbf{x}_1^{(0)},\ldots,\mathbf{x}_{N_a}^{(0)}]\bigr)
$$
where $\mathbf{x}_i^{(0)}$ is constructed using the learned interaction mask token $\mathbf{m}_\tau$ and no positional embedding. Removing positional priors ensures representations depend strictly on content and mutual context, treating the catalog as an unordered set.

\subsubsection{Pointer Decoder} 
The decoder constructs the ordering $\pi=[m_{\sigma(1)},\ldots,m_{\sigma(N_a)}]$ autoregressively using a 2-layer causal Transformer decoder. At step $t$,
  the decoder attends over the sequence of previously selected encoder representations $[\bar{\mathbf{h}}_{\sigma(1)},\ldots,\bar{\mathbf{h}}_{\sigma(t-1)}]$
  via causal self-attention, producing a decoder state $\mathbf{s}_t \in \mathbb{R}^d$. Each unselected asset $m_i$ is then scored via pointer attention:
  \[
  z_i^{(t)}=\mathbf{v}^{\top}\tanh(W_1\bar{\mathbf{h}}_i+W_2\mathbf{s}_t),
  \quad i\notin \mathrm{sel}_{<t},
  \]
  and selected according to
  \[
  p_\theta(m_{\sigma(t)}\mid \pi_{<t})
  =
  \mathrm{softmax}\left(\{z_i^{(t)}\}_{i\notin \mathrm{sel}_{<t}}\right),\textit{where } \mathbf{v}, W_1, W_2
  \]
   are learnable projections and $\mathbf{s}_1$ is a learnable intital state. Masked softmax over unselected assets guarantees a valid permutation. 

\subsection{Policy Optimization with Survival-Weighted Reward-to-Go}
\label{sec:policy_optimization}
Here, $U$ denotes the frozen content-based utility from Eq.~\eqref{eq:utility_head}, with the PAL tower removed. Generated orderings are scored in the same pre-serving setting as inference: in-session interactions are replaced by the learned mask token, so policy optimization uses only information available before carousel interaction. For a generated ordering $\pi=[m_{\sigma(1)},\ldots,m_{\sigma(N_a)}]$, we define the incremental utility of the asset placed at position $t$ as
$$
\Delta_t(\pi) = U(\pi_{\le t}) - U(\pi_{\le t-1}),
$$
where $U(\pi_{\le 0})$ is initialized to the empirical base conversion rate. This marginal formulation removes the constant base utility: an uninformative asset yields $\Delta_t \approx 0$, preventing the policy from accumulating reward merely from longer prefixes.

A natural objective is the expected utility of the prefix actually reached by the customer, $\mathbb{E}_{T}[U(\pi_{\le T})]$. Using cumulative marginal gains, this decomposes as
$$
\mathbb{E}_{T}[U(\pi_{\le T})]
=
U(\pi_{\le 0})
+
\sum_{t=1}^{N_a}
\Pr(T \ge t)\Delta_t(\pi).
$$
Thus, the marginal utility at position $t$ should be weighted by the survival probability of that position being reached, rather than by the probability of stopping exactly at $t$. 
Estimating $\hat{\rho}_t=\Pr(T_{\mathrm{obs}}\ge t)$ from logged session depths, we define the step reward as
\begin{equation}
r_t(\pi)
=
\hat{\rho}_t \gamma^{t-1}\Delta_t(\pi).
\end{equation}
where discount factor $\gamma\in(0,1]$ optionally emphasizes earlier persuasive evidence. Since $\hat{\rho}_t$ decreases with depth on mobile carousels, utility gains obtained at early positions receive larger reward than identical gains placed later. The policy is therefore explicitly optimized to surface conversion-informative media before likely customer drop-off, which directly targets reduced swipe depth.
We optimize the autoregressive policy $p_\theta$ using reward-to-go (RTG) REINFORCE~\cite{williams1992reinforce,sutton2018rl}. For a sampled ordering $\pi$, the return from position $t$ onward is:
\vspace{-2mm}
$$
G_t(\pi)=\sum_{j=t}^{N_a} r_j(\pi).
$$
Unlike terminal REINFORCE, which assigns the same slate reward to every action, RTG provides a temporally localized credit signal by assigning each decoding decision only its future rewards. To reduce variance, we use a self-critical sequence training (SCST) baseline \cite{rennie2017self}. For each product slate, we compute a greedy ordering $\pi^{\mathrm{gr}}$ using argmax decoding from the current policy and define
\vspace{-1mm}
$$
A_t(\pi)=G_t(\pi)-G_t(\pi^{\mathrm{gr}}).
$$
The greedy baseline is treated as stop-gradient and computed by deterministic decoding under the current policy. This yields a low-cost policy-dependent baseline requiring one additional rollout per batch, which is practical in our setting where $N_a$ is small.

The entropy-regularized policy loss is
$$
\mathcal{L}_{\pi}
=
-
\sum_{t=2}^{N_a}
\log p_\theta(m_{\sigma(t)}\mid\pi_{<t})
\,\mathrm{sg}\!\left(A_t^{(k)}\right)
-
\lambda_H
\sum_{t=2}^{N_a}
\mathcal{H}\!\left(p_\theta(\cdot\mid\pi_{<t})\right),
$$
where $\mathrm{sg}(\cdot)$ denotes the stop-gradient operator and $\mathcal{H}$ is the Shannon entropy. 
\textbf{Note:} The summation starts at $t=2$ because the primary product image is conventionally fixed at $t=1$, a common product-page UX constraint. SMEO therefore optimizes only the subsequent slots conditional on this fixed initial state, ensuring generated orderings respect layout conventions. The entropy bonus ($\lambda_H=5\times10^{-4}$) prevents premature policy collapse while offline advantages are noisy. The pointer network restricts actions to valid product media and masks already selected assets, preventing invalid or duplicate outputs. We tune $\lambda_H$ and monitor held-out metrics to detect potential reward over-optimization.

\subsection{Media Contribution Estimation}
\label{sec:attribution}

The frozen content-based utility model also enables offline media diagnostics. Here, $U$ denotes the utility model with the PAL tower removed and in-session interactions masked. Given ordering $\pi$, we estimate media asset ($m_j^a$) contribution via leave-one-out masking:
\begin{equation}
  \varphi_{\mathrm{LOO}}(m_j^a \mid \pi)
  =
  U(\pi) - U(\pi \setminus \{m_j^a\}),
  \label{eq:loo}
\end{equation}
where $\pi \setminus \{m_j^a\}$ removes $m_j^a$ and re-indexes the remaining assets. This context- and ordering-dependent score helps identify high value or redundant media and whether specific media types should be surfaced earlier.  We also compute evidence accumulation curves, $E_t(\pi)=U(\pi_{\le t})$, to visualize how quickly an ordering builds conversion utility before likely drop-off. These diagnostics, presented in Section~\ref{sec:rq3_diagnostics}, provide category audit and merchandising teams in e-commerce with interpretable signals for prioritizing, replacing, or requesting media assets. These are interpretability diagnostics; primary evaluation uses observed conversion-oriented metrics.

\section{Experiments}
\label{sec:experiment}
We investigate the following research questions:
\begin{itemize}[leftmargin=*]
    \item \textbf{RQ1:} Does SMEO improve conversion and swipe efficiency?
    \item \textbf{RQ2:} Which SMEO components drive the gains?
    \item \textbf{RQ3:} Does SMEO learn interpretable media-ordering behavior?
\end{itemize}

\subsection{Experimental Setup}

\textbf{Dataset:} We utilize a large-scale, proprietary dataset comprising of anonymized mobile session logs from a global e-commerce store
  spanning several weeks across multiple product categories (electronics, apparel, home, beauty): approximately 150M randomly sampled sessions over millions of unique products with tens of millions of total media assets (static images, videos, and 3D interactive renders). Each session records the product identifier, ordered media slate, per-asset
  interaction signals (dwell time, click and zoom counts), observed stopping depth $T^s_{\mathrm{obs}}$, and terminal binary purchase outcome $y^s$. We
  evaluate on a temporally held-out test set to prevent data leakage. Candidate slates are capped at $N_{\max}=15$.

\noindent \textbf{Experimental Details: }  All models are trained on 8$\times$ NVIDIA V100 GPUs with frozen, precomputed visual encoders. The trainable
  architecture—multimodal projections, trajectory utility model, PAL tower, and pointer policy—totals $\sim$4.5M parameters. The trajectory encoder uses
  $L{=}4$ causal Transformer layers ($H{=}4$ heads, $d{=}256$); the pointer policy uses a 2-layer autoregressive decoder with identical dimensionality. We optimize with AdamW, lr= $10^{-4}$, 10\% warmup, and cosine decay. The utility
  model trains for 15 epochs with early stopping on validation BCE. The pointer policy subsequently trains for 10 epochs using the frozen $U$ as a dense reward
   signal. To stabilize policy optimization, we apply per-batch advantage normalization, 
  survival discount $\gamma{=}0.90$, entropy regularisation $\lambda_H = 5{\times}10^{-4}$ and warm-start the decoder via supervised pre-training on logged historical orderings for 2 epochs.

\noindent \textbf{Baselines: }We compare SMEO against five baselines:
\begin{enumerate}[leftmargin=*]
    \item \textbf{Heuristic:} a rule-based ordering that serves as our real-world anchor baseline.

    \item \textbf{CTR:} a pointwise model that optimizes on per asset engagement proxy (clicks) rather than terminal conversion.

    \item \textbf{CLTR-CVR:} a pointwise conversion model trained with standard Counterfactual LTR \cite{joachims2017unbiased}. It assigns the terminal session outcome $y^s$ to each viewed asset independently and corrects for position bias using Inverse Propensity weighting. This tests whether item-level debiasing suffices without sequential inter-asset context, while retaining the conversion objective.

    \item \textbf{Listwise-MM:} a multimodal listwise ranker \cite{ai2018learning,pei2019personalized} that predicts terminal conversion from a self-attended slate representation. This retains inter-asset context but does not model consumed-prefix utility or autoregressive decoding.

    \item \textbf{Seq2Slate-Adapt:} an autoregressive pointer network \cite{bello2019seq2slate} with the same policy architecture as SMEO, trained using terminal slate-level REINFORCE on the session outcome
    $y^s$ with an SCST baseline \cite{rennie2017self}. This isolates the value of SMEO's consumed-prefix utility learning and survival-weighted RTG objective.
\end{enumerate}
All learned baselines use the same features as SMEO, so differences reflect modeling choices rather than feature access.

\noindent

\noindent \textbf{Metrics.}
We evaluate offline conversion value and early-swipe efficiency. The terminal label $y^s\in\{0,1\}$ denotes whether session $s$ ended in purchase. Although logged ordering is not randomized, it varies across categories due to overlapping business rules, catalog updates, and upstream experiments. In our data, a large majority of products have multiple historical media permutations, providing partial support for off-policy evaluation. We therefore restrict estimates to supported regions and use clipping to control variance.

\textbf{SNIPS-CVR.}
We report clipped self-normalized IPS-CVR~\cite{swaminathan2015self}:
\[
\widehat{\mathrm{CVR}}_{\mathrm{SNIPS}}(\pi)
=
\frac{\sum_s \bar{w}_s y^s}{\sum_s \bar{w}_s},
\qquad
\bar{w}_s
=
\min\!\left(
\frac{\hat p_{\pi}(\pi^s \mid a_s)}
     {\hat p_{\mu}(\pi^s \mid a_s)},
W_{\mathrm{IPS}}
\right),
\]
Here, $\pi^s$ denotes the logged ordering in session $s$, $\pi(a_s)$ denotes the ordering induced by the evaluated target policy $\pi$ for product $a_s$.
$\hat p_\mu$ is a behavior-cloned logging model, and $\hat p_\pi$ is the stochastic distribution induced by the target policy. For deterministic rankers, $\hat p_\pi$ is computed using a Plackett--Luce distribution over model scores~\cite{plackett1975analysis}; pointwise models sort by score, Listwise-MM greedily decodes marginal gain, and pointer models decode autoregressively.


\textbf{DR-CVR.}
We also report doubly robust CVR~\cite{dudik2011doubly,swaminathan2015counterfactual}:
\[
\widehat{\mathrm{CVR}}_{\mathrm{DR}}(\pi)
=
\frac{1}{|\mathcal{D}|}
\sum_s
\left[
\hat{r}(a_s,\pi(a_s))
+
\bar{w}_s
\left(y^s-\hat{r}(a_s,\pi^s)\right)
\right],
\]
where $\hat{r}(a,\pi)$ is a $K$-fold cross-fitted predictor of $\mathbb{E}[y\mid a,\pi]$ from product features and media ordering. DR-CVR combines direct reward prediction with logged-policy correction.

\textbf{MSC.}
For converted sessions, let $r_i^s=\mathbb{I}\{m_i \text{ is engaged in }s,\; y^s=1\}$. For ordering $\pi(a_s)$, where $\pi_k(a_s)$ is the asset placed at rank $k$:
\[
\mathrm{MSC}
=
\mathrm{median}_{s:y^s=1}
\left[
\min\{k:r_{\pi_k(a_s)}^s=1\}
\right].
\]
MSC measures how early a policy surfaces historically conversion-associated media; lower is better. Since stopping behavior may change under a new ordering, MSC is not a causal swipe estimate but a relative ordering-quality metric for early evidence placement.

\subsection{RQ1: SMEO vs. Baselines}
Table~\ref{tab:main_results} shows three trends. \textbf{First, engagement proxies are insufficient:} optimizing CTR yields only marginal conversion lift and almost no swipe-efficiency improvement, indicating that attention-grabbing media does not necessarily provide purchase-decisive evidence. \textbf{Second, conversion objective and context both matter:} CLTR-CVR improves DR-CVR over engagement ranking, but remains limited by independent asset scoring and unmodeled sequence redundancy. Listwise-MM adds inter-asset context, producing an additional $+1.6$ point relative DR-CVR gain over CLTR-CVR and a larger MSC reduction. \textbf{Finally, trajectory-aware reward design is most effective:} SMEO improves over Seq2Slate-Adapt by $+1.9$ points in relative DR-CVR and $4.5$ points in MSC reduction. While both use autoregressive decoding, SMEO improves beyond Seq2Slate-Adapt by restructuring the reward signal: survival-weighted RTG credits marginal utility gained before likely drop-off, directly optimizing early evidence delivery.

\begin{table}[t]
\centering
\small
\caption{For confidentiality, all metrics are reported as relative change against the heuristic baseline. Values include 95\% confidence intervals. Higher is better for SNIPS-CVR, DR-CVR; lower is better for MSC.}
\label{tab:main_results}
\begin{tabular}{lccc}
\toprule
\textbf{Method} 
& \textbf{$\Delta$SNIPS-CVR $\uparrow$} 
& \textbf{$\Delta$DR-CVR $\uparrow$} 
& \textbf{$\Delta$MSC $\downarrow$} \\
\midrule
Heuristic (Rule-based)        & -- & -- & -- \\
Engagement(CTR)  & $+0.7{\pm}0.5\%$ & $+0.4{\pm}0.3\%$ & $-0.8{\pm}0.5\%$ \\
CLTR-CVR          & $+1.5{\pm}0.7\%$ & $+1.1{\pm}0.5\%$ & $-3.6{\pm}0.8\%$ \\
Listwise-MM       & $+3.2{\pm}0.6\%$ & $+2.7{\pm}0.4\%$ & $-8.5{\pm}1.2\%$ \\
Seq2Slate-Adapt   & $+4.1{\pm}0.5\%$ & $+3.5{\pm}0.4\%$ & $-10.8{\pm}1.3\%$ \\
\textbf{SMEO}     & $\mathbf{+6.1{\pm}0.6\%}$ & $\mathbf{+5.4{\pm}0.3\%}$ & $\mathbf{-15.3{\pm}1.5\%}$ \\
\bottomrule
\end{tabular}
\end{table}

\begin{figure*}[t]
    \centering
    \begin{minipage}[t]{0.69\textwidth}
        \centering
        \includegraphics[width=\linewidth]{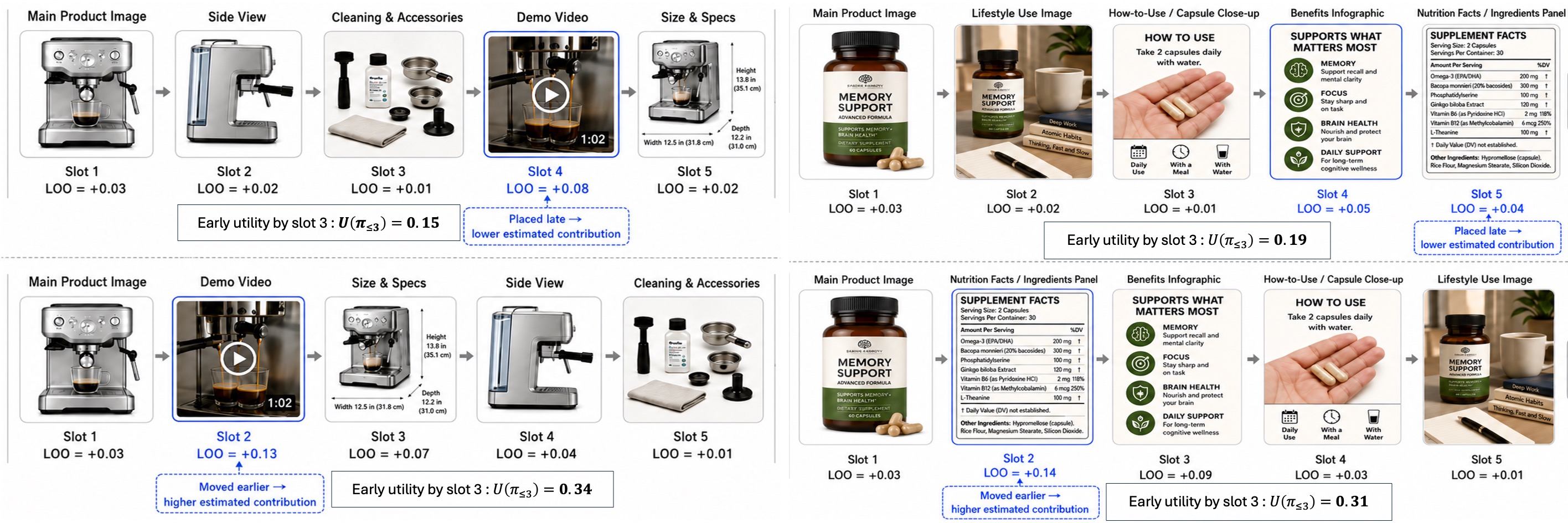}
        \vspace{-1mm}
        \centerline{\small \textbf{(a) Early Evidence Contribution:} (top) Heuristic baseline ranking vs (bottom) SMEO.}
    \end{minipage}
    \hfill
    \begin{minipage}[t]{0.29\textwidth}
        \centering
        \includegraphics[width=\linewidth,height=0.18\textheight]{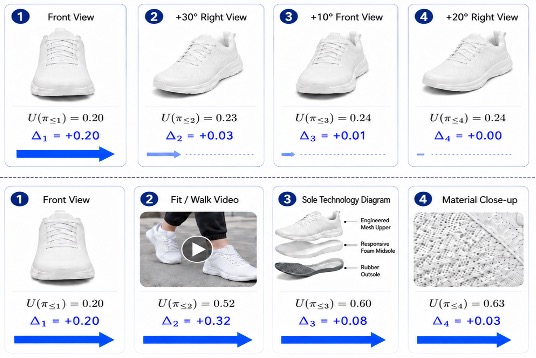}
        \vspace{-1mm}
        \centerline{\small \textbf{(b) Contextual Redundancy}}
    \end{minipage}
    \vspace{-1mm}
    \caption{(a) With the main product image fixed at slot 1, SMEO moves high-information assets earlier. Demo videos for coffee makers and nutrition labels for supplements produce larger LOO contribution when surfaced near the front. (b) For sneakers, near-identical views drive $\Delta_t = U(\pi_{\leq t}) - U(\pi_{\leq t-1})$ toward zero after the first image, while diverse assets sustain marginal utility by adding complementary evidence.}
    \label{fig:qualitative}
    \vspace{-2mm}
\end{figure*}

\subsection{RQ2: Ablation Study}
\label{sec:rq2_ablation}

We ablate SMEO-specific components in Table~\ref{tab:ablation} and report relative changes against full SMEO using the same offline evaluation protocol.

\begin{table}[t]
\centering
\small
\caption{Ablation study of SMEO components. Values are relative to full SMEO.}
\label{tab:ablation}
\begin{tabular}{lccc}
\toprule
\textbf{Variant} 
& \textbf{$\Delta$SNIPS-CVR $\uparrow$} 
& \textbf{$\Delta$DR-CVR $\uparrow$} 
& \textbf{$\Delta$MSC $\downarrow$} \\
\midrule
SMEO full              & -- & -- & -- \\
w/o PAL                & $-1.2{\pm}0.4\%$ & $-1.0{\pm}0.3\%$ & $+2.1{\pm}0.7\%$ \\
PAL tower kept in $U$  & $-1.8{\pm}0.5\%$ & $-1.5{\pm}0.4\%$ & $+3.4{\pm}0.8\%$ \\
w/o survival weighting & $-2.3{\pm}0.5\%$ & $-1.9{\pm}0.3\%$ & $+4.8{\pm}0.9\%$ \\
w/o interaction masking& $-0.8{\pm}0.4\%$ & $-0.7{\pm}0.3\%$ & $+1.5{\pm}0.6\%$ \\
\bottomrule
\end{tabular}
\end{table}

\noindent \textbf{Effect of Position-Aware Learning:}
Removing the PAL tower during Stage 1 training (\emph{w/o PAL}) forces the utility model to absorb historical position effects into the content encoder, reducing downstream DR-CVR by $1.0\%$ relative to full SMEO.

\noindent \textbf{Importance of Removing PAL at Policy Time:}
Training with PAL but keeping the PAL tower during Stage 2 policy optimization (\emph{PAL tower kept in $U$}) further worsens swipe efficiency ($\Delta\mathrm{MSC}=+3.4\%$). This suggests that exposing the policy to position-biased reward components encourages it to exploit learned slot effects rather than optimize content-driven sequencing.

\noindent \textbf{Survival Weighting Drives Early Placement:}
Removing survival weighting causes the largest MSC degradation, supporting customer reach probability as the key mechanism for front-loading persuasive evidence. This also helps explain SMEO's advantage over Seq2Slate-Adapt: autoregressive decoding alone is insufficient without reach-weighted marginal utility.

\noindent \textbf{Train--Serve Robustness:}
Removing in-session interaction masking (\emph{w/o interaction masking}) slightly degrades CVR and swipe efficiency. This indicates that masking improves robustness under page-load inference, where real-time interaction features are unavailable.

\begin{figure}[t]
    \centering
    \includegraphics[width=\linewidth]{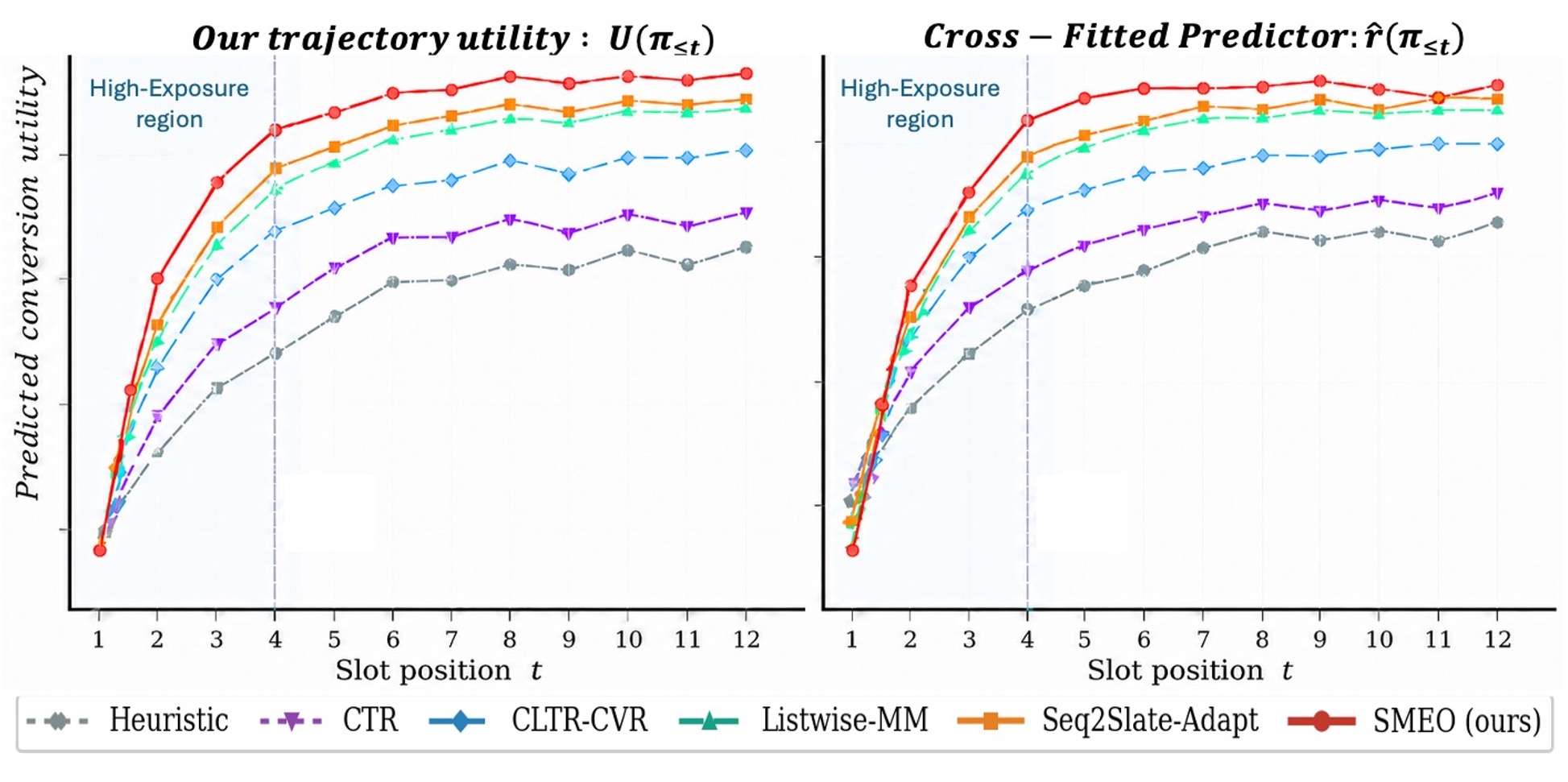}
    \vspace{-7mm}
    \caption{Evidence accumulation curves across prefix depth.}
    \label{fig:utility_curves}
    \vspace{-3mm}
\end{figure}

\subsection{RQ3: Qualitative Media Diagnostics}
\label{sec:rq3_diagnostics}

\textbf{Evidence Front-Loading:}
Figure~\ref{fig:utility_curves} plots prefix-level predicted conversion utility as media are revealed slot by slot. In the left panel, this is $E_t(\pi)=U(\pi_{\le t})$, the trajectory utility score for the first $t$ assets; SMEO rises fastest, as expected since it optimizes $U$. To avoid circular evaluation, the right panel scores the same prefixes with the independent cross-fitted predictor $\hat r(a,\pi_{\le t})$ from Sec.~\ref{sec:experiment}. SMEO still leads through the early slots, 
before likely customer drop-off, while Listwise-MM and Seq2Slate-Adapt partially catch up later. This indicates that SMEO primarily improves \emph{earlier} placement of conversion-informative media rather than merely increasing full-slate utility. Fig.~\ref{fig:qualitative}(a) provides qualitative evidence: with slot 1 fixed by the standard product-page convention, moving the same high-information asset earlier yields larger LOO (Eqn. \ref{eq:loo}) contribution and higher $U(\pi_{\le 3})$, showing that ordering affects both marginal media value and early evidence accumulation.

\vspace{5pt}
\noindent\textbf{Redundancy correlation:}
We quantify redundancy by correlating each asset's marginal gain,
$\Delta_t = U(\pi_{\le t}) - U(\pi_{\le t-1})$,
with its maximum visual similarity to the preceding prefix,
$
s_t^{\max}
=
\max_{k<t}
\cos(\mathbf{v}_{\sigma(t)}, \mathbf{v}_{\sigma(k)}),
$
where $\mathbf{v}_{\sigma(t)}$ is the visual embedding of the asset at position $t$. To remove position-depth confounding, we compute partial Spearman correlation controlling for $t$. Strong negative correlation ({\boldmath $\rho_{\mathrm{partial}}=-0.61$, $p<0.001$}) shows that visually redundant assets contribute less incremental utility (Fig. \ref{fig:qualitative}(b)).

\vspace{5pt}
\noindent\textbf{Category-Level Attribution Patterns:}
Aggregating $\varphi_{\mathrm{LOO}}$ by modality and category shows that SMEO learns category-specific media roles that are meaningful for e-commerce merchandising. \textbf{3D assets contribute significantly more in spatial categories such as furniture and home fixtures} than elsewhere ($0.091{\pm}0.019$ vs.\ $0.028{\pm}0.011$, $p{<}0.01$), where customers need geometry, scale, and room-fit evidence. \textbf{Videos dominate dynamic categories such as appliances and activewear} ($0.103{\pm}0.024$ vs.\ $0.041{\pm}0.016$, $p{<}0.001$), where operation, motion, or fit-in-use must be demonstrated. \textbf{Specification and ingredient images contribute most in high-scrutiny categories such as electronics and consumables} ($0.078{\pm}0.017$ vs.\ $0.022{\pm}0.009$, $p{<}0.01$), where factual details reduce purchase uncertainty. This confirms that  SMEO learns media-type-specific persuasion roles from conversion signal alone,without any supervision.
This enables offline decisions about which assets to prioritize, audit, or request for each product category, and provides actionable guidance for seller content improvement in e-commerce.

\section{Batch Inference and Serving}
\label{sec:deployment}

SMEO is designed as an offline ranking system decoupled from real-time page serving. In a representative deployment configuration, media orderings are generated from static product-level signals ---visual embeddings, historical priors, and product/media metadata---with no dependence on real-time customer state at page load. SMEO uses a distributed batch pipeline with precomputed embeddings and cached features. The short decode length ($N_{\max}{=}15$) makes generation tractable: in our experiments, a full refresh on a 16-GPU V100 cluster completes in a few hours, with inference accounting for less than half of wall-clock time. The pipeline scales horizontally — refresh cost is linear in catalogue size — and supports catalogue-scale operation through periodic incremental refreshes.
Generated orderings are written to a distributed key-value store keyed by product identifier and retrieved at request time through a single point lookup (sub-5 ms latency). When an ordering is unavailable, the system falls back to a default ordering, preserving availability. 

\section{Conclusion}
We introduced SMEO, a two-stage framework that treats product media as sequential, cooperative evidence for purchase decisions. By learning consumed-prefix conversion utility and optimizing a survival-weighted autoregressive policy, SMEO surfaces the most decision-relevant media early, before likely customer drop-off. Offline evaluation shows consistent gains over heuristic, pointwise, listwise, and autoregressive baselines in estimated conversion and swipe efficiency. Diagnostics further reveal redundancy-aware gains and category-specific media contributions. Online exploration and personalization are natural extensions.

\clearpage

\section{Gen-AI Usage Disclosure}
In preparing this manuscript, generative AI tools were used solely
for language refinement—including improving grammar, style, clar-
ity, and readability of text that was originally authored by the pa-
per’s authors. No sections of the manuscript, including the abstract,
related work, methodology, experimental results, or conclusions,
were generated by AI systems. All conceptual contributions, techni-
cal content, experimental design, data analysis, and interpretations
are the authors’ own original work. The authors accept full respon-
sibility for the correctness and integrity of the manuscript.
\bibliographystyle{ACM-Reference-Format}
\balance
\bibliography{sample-base}

\end{document}